\documentclass[runningheads]{llncs}

\usepackage{graphicx}
\usepackage{subfigure}
\usepackage{amsmath,amssymb,amsfonts}
\usepackage{color}
\usepackage{ulem}
\usepackage{hyperref}
\usepackage{marvosym}
\hypersetup{
    colorlinks=true,
    linkcolor=blue,
    filecolor=magenta,      
    urlcolor=cyan,
    pdftitle={Overleaf Example},
    pdfpagemode=FullScreen,
    }
\makeatletter
\newcommand{\printfnsymbol}[1]{%
  \textsuperscript{\@fnsymbol{#1}}%
}
\makeatother

\begin{document}
\title{Neural State-Space Modeling with \\Latent Causal-Effect Disentanglement}
\titlerunning{Latent Causal-Effect Modeling}

\author{Maryam Toloubidokhti\inst{1}\thanks{Both authors contributed equally to this work \\ Corresponding author: Maryam Toloubidokhti, mt6129@rit.edu \\\\\\ 13th Machine Learning in Medical Imaging (MLMI 2022) Workshop.}  \and
Ryan Missel\inst{1}\printfnsymbol{1} \and
Xiajun Jiang\inst{1} \and
\\Niels Otani\inst{1} \and
Linwei Wang\inst{1}}

\authorrunning{M. Toloubidokhti et al.}

\institute{
Rochester Institute of Technology, Rochester NY 14623, USA \\
\email{\{mt6129, rxm7244\}@rit.edu}\\
}

\maketitle 

\begin{abstract} 
Despite substantial progress in deep learning approaches to time-series reconstruction, no existing methods are designed to uncover local activities with minute signal strength due to their negligible contribution to the optimization loss. Such local activities however can signify important abnormal events in physiological systems, such as an extra foci triggering an abnormal propagation of electrical waves in the heart. We discuss a novel technique for reconstructing such local activity that, while small in signal strength, is the cause of subsequent global activities that have larger signal strength. Our central innovation is to approach this by explicitly modeling and disentangling how the latent state of a system is influenced by potential \textit{hidden internal interventions.} In a novel neural formulation of state-space models (SSMs), we first introduce causal-effect modeling of the latent dynamics via a system of interacting neural ODEs that separately describes 1) the continuous-time dynamics of the internal intervention, and 2) its effect on the trajectory of the system's native state. Because the intervention can not be directly observed but have to be disentangled from the observed subsequent effect, we integrate knowledge of the native \textit{intervention-free} dynamics of a system, and infer the hidden intervention by assuming it to be responsible for differences observed between the actual and hypothetical \textit{intervention-free} dynamics. We demonstrated a proof-of-concept of the presented framework on reconstructing ectopic foci disrupting the course of normal cardiac electrical propagation from remote observations.

\end{abstract}

\section{Introduction}
Advancements in modern deep learning have resulted in substantial progress in our ability to model, reconstruct, and forecast complex time-series data, most notably through recurrent neural networks (RNNs) and neural ordinary differential equations (neural ODEs) \cite{chen2019neuralODE,rubanova2019latentODE,rangapuram2018deep,li2021learning}. Many of these techniques have been successfully
adapted to model the spatiotemporal dynamics of complex physiological systems, such as in reconstructing \cite{jiang2021label,ghimire2018generative} or forecasting the electrical activity of the cardiac system \cite{EPNET2021}. 
While strong in modeling general system dynamics, however, these methods are not designed to uncover local activity with minute signal strength compared to the global dynamics. Such local activities are common and often signify important abnormal events in physiological systems, such as an extra foci that triggers an abnormal propagation of electrical waves amid normal propagation patterns (Figure \ref{fig:normalProp}). 
Due to negligible optimization loss contribution given weak signal strength, reconstructing these local activities remains challenging despite deep learning advances in time-series modelling.

We present a fundamentally novel perspective of this problem: instead of general reconstruction, we stress that these local activities, while small in signal strength, are the cause of the subsequent global activities that have larger signal strength. Therefore, 
we may be able to disentangle the \textit{cause} and \textit{effect} of the latent dynamics 
by explicitly modeling and inferring how the latent state of a system is influenced by potential hidden internal \textit{interventions}. 
Modeling the effect on system dynamics by exogenous inputs is well-studied in classic state-space systems (SSMs), and has seen recent successes in neural-network modeling of SSMs \cite{krishnan2015deep,de2019gruODEBayes,karl2016deep,gwak2020imode}. 
These works, however, focus on external interventions that are either known or directly observed, whereas we consider hidden and internal causes of system's abnormal events. 
As these internal causes are not directly observable, we propose to disentangle them from the observed effects on the system's dynamics. Rather than learning the complex intervened dynamics of a system from scratch as in existing works \cite{gwak2020imode}, we incorporate knowledge of the native \textit{intervention-free} dynamics of a system and focus only on learning how it is influenced by potential hidden interventions.

To this end we present a novel neural SSM with two major innovations. First, drawing inspirations from related work in intervention modeling \cite{gwak2020imode}, we introduce causal-effect modeling of the latent dynamics via a system of interacting neural ODEs that separately describe 1) the continuous-time dynamics of the internal intervention, and 2) its effect on the trajectory of the system's native state. Second, to disentangle the cause and effect from their collective observations, we 1) leverage data of intervention-free systems to pre-train a neural ODE of \textit{native} dynamics and  integrate it into the intervention ODEs, and 2) infer the hidden intervention by assuming it to be responsible for differences observed between the actual and hypothetical \textit{intervention-free} dynamics at each time frame.

We demonstrated a proof-of-concept of the presented framework on reconstructing ectopic foci that disrupt normal electrical propagation using remote observations. We compared the presented method to the two most relevant time-series modeling approaches: 
1) a global neural ODE to describe latent system dynamics, 
and 2) a neural SSM with a latent neural ODE lacking intervention modeling, developed for reconstructing cardiac electrical propagation \cite{jiang2021label}.  
Experiments were conducted on synthetic data with controlled \textit{internal interventions}, simulated in 2D settings.
Our results indicated that the presented method delivers more accurate inverse estimations in terms of localizing the triggering events. 

\section{Methodology}

We formulate a neural SSM where the system dynamics is described on a lower-dimensional latent manifold in separation from its emission to the data space as illustrated in Fig.~\ref{fig:overview}.
Consider cardiac electrical propagation $\mathbf{x_t}$ and its body-surface measurement $\mathbf{Y_t}$ with a physics-based relation $\mathbf{Y_t} = \mathbf{H}\mathbf{x}_t$.
We enable causal-effect intervention modelling by sequentially learning two separate neural ODE functions, $\mathcal{F}_z$ and $\mathcal{F}_a$, that describe \textit{native} system dynamics and causal dynamics, respectively. The effect of intervention $\mathbf{a}$ on $\mathbf{z}$ is modelled through a coupled neural ODE function, $\mathcal{F}_{z, a}$. 
Because intervention $\mathbf{a}$ is not directly observable, 
it is not possible to separately learn $\mathcal{F}_z$, $\mathcal{F}_a$ and $\mathcal{F}_{z, a}$ all from scratch from only their collective observations.
Instead, we poise that we must leverage knowledge of native dynamics $\mathcal{F}_z$, in order to 
be able to generate \textit{hypothetical observation} 
$\hat{\mathbf{Y}}_t = \mathbf{H} \hat{\mathbf{x}}_t$ 
assuming intervention-free dynamics since last observation $\mathbf{Y}_{t-1}$. Then using the residual between the actual observation $\mathbf{Y}_t$ and hypothetical observation
$\hat{\mathbf{Y}}_t$ allows us to model and estimate intervention $\mathbf{a}_t$. 

A two-stage optimization process is leveraged: First, the \textit{native} dynamics of the system are learned on an \textit{intervention-free} subset of data (termed \textit{ODE-VAE}) which is then statically integrated into a system of intervention ODEs optimized on the intervention set (termed \textit{ODE-VAE-IM}).

\begin{figure}[!tb]
\label{fig:overview}
    \begin{center}
        \includegraphics[width=\textwidth]{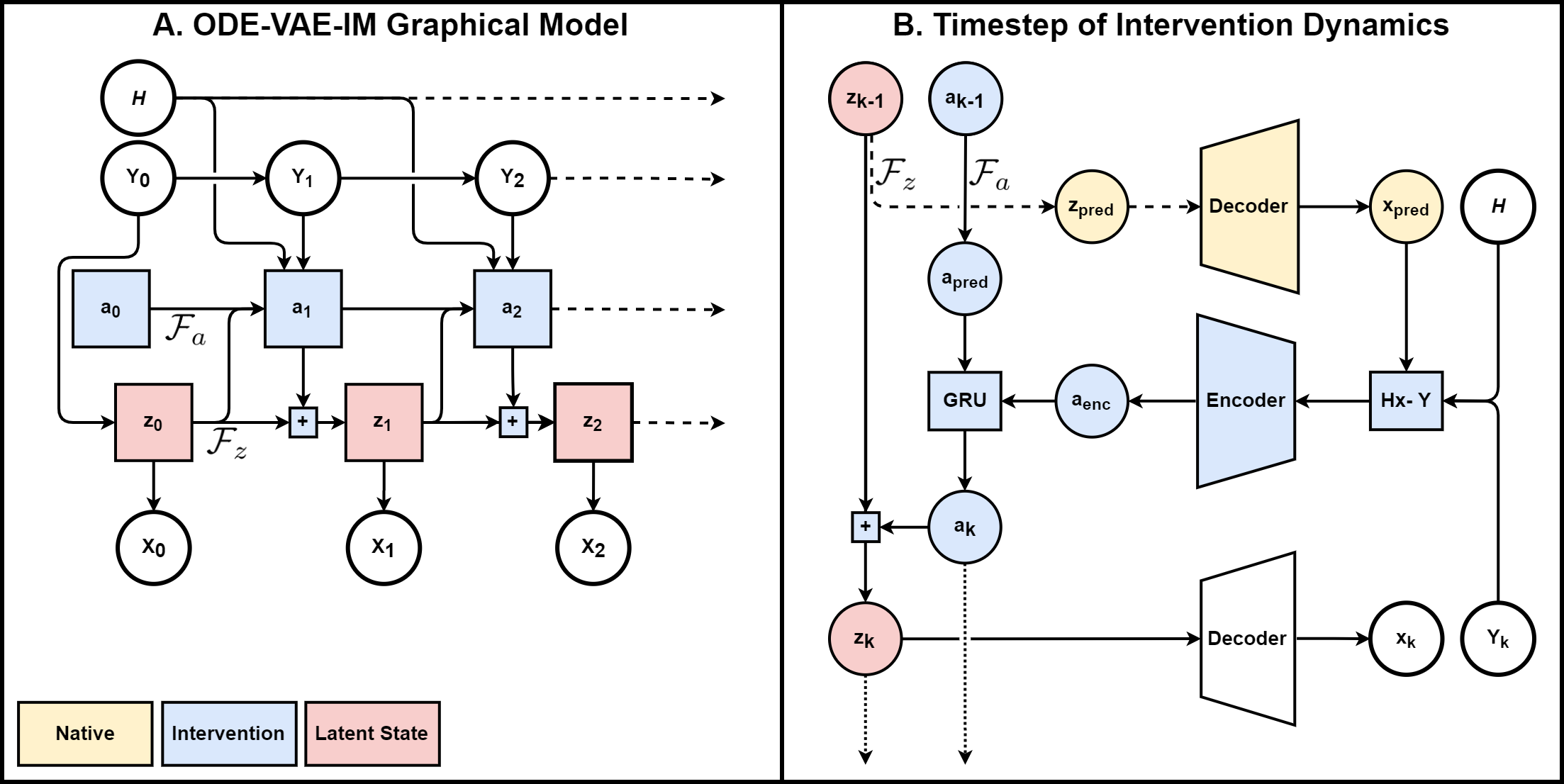}
        \caption{A) Graphical overview of the proposed network. Inputs \textbf{H} and $\mathbf{Y}_{0:T}$ are given to the intervention dynamics $\mathcal{F}_{a}$ and influence the propagation of $\mathcal{F}_{(z,a)}$. B) Schematic of a single intervention step, showing the estimation of the intervention latent variable $\mathbf{a}_k$ from data-space observations and \textit{hypothetical intervention-free observations}.
        }
        \label{fig:overview}
    \end{center}
\end{figure}

In sections \ref{sec:naiveDynamics}-\ref{sec:dynamics} we describe the native dynamics model, the predictive model enabling causal-effect learning, the estimation of the latent intervention variable through global observations, and the model optimization loop.

\subsection{Modeling and Learning Intervention-Free Native Dynamics} \label{sec:naiveDynamics}
An ODE function $\mathcal{F}_z$ is designed to learn the \textit{native} dynamics function that handles the propagation of the system in the absence of interventions. Since we leverage this function to hypothesize the intervention-free native state of a system given at any previous system state, it must have a strong long-term forecasting ability. Therefore, inspired by \cite{yildiz2019ode2vae}, 1.) an encoding network $Enc_z$ initializes the latent ODEs vector field $\mathbf{z}_0$ using the first few frames of sequential input $\mathbf{Y}_{0:k}$ (Eq.\ref{eqn:z_ode_1}), 2.) an initial value problem is solved up to time $\mathbf{T}$ using the ODE function $\mathcal{F}_z$ (Eq.\ref{eqn:z_ode_2}), and 3.) a decoding network $Dec_z$ converts the latent trajectory $\mathbf{z}_{0:T}$ to the output space $\mathbf{X}_{0:T}$ (Eq.\ref{eqn:z_ode_3}): 
\begin{align}
    \mathbf{z}_0 &= Enc_z(\mathbf{Y}_{0:k})
    \label{eqn:z_ode_1} \\
    \mathbf{z}_i &= \mathbf{z}_{i-1} + \int_{t_{i-1}}^{t_i} \mathcal{F}_{z}(\mathbf{z}_\tau) d\tau
    \label{eqn:z_ode_2} \\
    \hat{\mathbf{x}}_i &= Dec_z(\mathbf{z}_i),
    \label{eqn:z_ode_3}
\end{align}
where $Enc_z$ and $Dec_z$ are represented by CNNs and the sequence input length $k$ is empirically tuned. 

We train this native dynamics model on datasets of \textit{intervention-free} systems.  
After initializing the latent state, the full trajectory is predicted using $\mathcal{F}_{z}$, from which the reconstructed $\hat{\mathbf{x}}$
is compared with the ground truth in the loss: 
\begin{align}
    \mathcal{L}_{native}(\mathbf{x}_{0:T}, \hat{\mathbf{x}}_{0:T}) &= 
    \beta [(\mathbf{x}_0 * log(\hat{\mathbf{x}}_0)) + (1 - \hat{\mathbf{x}}_0) * log(1 - \hat{\mathbf{x}}_0)] \\
    &+ \frac{1}{T} \sum_{i=1}^T (\mathbf{x}_i * log(\hat{\mathbf{x}}_i) + (1 - \hat{\mathbf{x}_i}) * log(1 - \hat{\mathbf{x}}_i),
\end{align}
where binary cross entropy (BCE) is applied to each frame and $\beta$ represents a loss weighting coefficient. We explicitly separate out the reconstruction of $\mathbf{x}_0$ to emphasize the accuracy of the initial conditions, $\mathbf{z}_0$, independent of the accuracy of the dynamic transition models. Because the network is asked to reconstruct the full sequence of $\mathbf{x}_{0:T}$ from only an estimated initial condition $\mathbf{z}_0$, it promotes learning strong, continuous dynamics models $\mathcal{F}_{z}$ that can simulate the system given any initial condition of the space. 
This function, once trained, will be incorporated into the intervention-effects function $\mathcal{F}_{(z, a)}$ as a static dynamics function whose trajectory is influenced by the intervention state.

\subsection{Modelling and Learning Intervention Dynamics} \label{sec:dynamics}

The system dynamics predictive model is described by two ODE functions, $\mathcal{F}_a$ and $\mathcal{F}_{(z, a)}$, that aim to separately estimate the cause and effect of interventions in the latent space, respectively. To achieve this, we use the state-space Bayesian filtering setting where the system propagation undergoes a series of predict-from-dynamics and update-from-observations steps \cite{karl2016deep,fraccaro2017disentangled,de2019gruODEBayes}. The former comprises of the predictive model which, given a current state $\mathbf{s}_{t-1}$, predicts the next temporal state, $\hat{\mathbf{s}}_{t}$, using the transition function $\mathcal{F}$. The latter corrects the temporal prediction through an update function, $\mathcal{G}$, using an inferred state $\mathbf{s}_{enc}(t)$ from the current observation $\mathbf{Y}_i$.

\textbf{Predictive System Dynamics Model.}
$\mathcal{F}_a$ handles the prediction of the intervention state, $a_t$, while $\mathcal{F}_{(z, a)}$ handles the intervention-influenced prediction of the latent state $z_t$, denoted by the equations in Eq.~\ref{eqn:combined}, respectively. 
\begin{align}
    \mbox{\textbf{Prediction:} }
    \begin{bmatrix}
        \hat{\mathbf{a}}_i \\
        \hat{\mathbf{z}}_i 
    \end{bmatrix}
    &=
    \begin{bmatrix}
        \mathbf{a}_{i-1} \\
        \mathbf{z}_{i-1} 
    \end{bmatrix}
    + \int_{t_{i-1}}^{t_i}
    \begin{bmatrix}
        \mathcal{F}_{a}(\mathbf{a}_\tau) \\
        \mathcal{F}_{z}(\mathbf{z}_\tau) + \mathcal{F}_{a}(\mathbf{a}_\tau)
    \end{bmatrix}
    d\tau \label{eqn:combined}
\end{align}

\textbf{Estimating the Intervention Latent Variable.}
Different from Section 2.1 wherein our goal is to learn a native dynamic model that has strong standalone forecasting ability, 
here our interest is to extract information from available observations at each time frame to uncover hidden interventions. 
A dynamical function describing the temporal cause of intervention alone is not enough to produce a meaningful causal-effect model. Rather, an observational source from which the latest intervention information can be inferred is required \cite{gwak2020imode,yin2021augmentingaphyn}. We hypothesize that the latent variable associated with the interventions rises from differences observed between data space observations and \textit{hypothetical observations} continuing under \textit{intervention-free} dynamics at each timestep. For inverse image reconstruction problems, this can be formulated as the element-wise difference in the inverse solution difference of $H\mathbf{\hat{x}}^\mathbf{z}_t$ and $\mathbf{Y}_t$. This difference is then used as input to an encoder to output the estimated latent intervention state.
\begin{align}
    \mbox{\textbf{Update:} } \mathbf{a}_{enc} &= Enc_a(|| \mathbf{H}\hat{\mathbf{x}_i} - \mathbf{Y}_i ||_{i:(i+k)}) \label{eqn:pienc} \\
    \mathbf{a}_i &= \mathcal{G}_a(\hat{\mathbf{a}}_{i}, \mathbf{a}_{enc}) \label{eqn:z_ode}
\end{align}
We make use of the Gated Recurrent Unit (GRU) cell \cite{cho2014learningGRU}, $\mathcal{G}_a$ to learn a weighted combination of the encoded intervention variable and the prediction coming from the intervention dynamics function.
A number of methods may be applied to improve the forward intervention estimation. One such choice, and the one utilized in this work, is to predict a number of timesteps ahead and get the inverse solution difference over a sequence of frames rather than just the current predicted frame. This provides increased temporal information to the intervention encoder by exposing the longer-term differences that would arise from the current state.

\textbf{Optimizing \textit{Intervention} Models.}
 The loss function is the  data reconstruction loss averaged over each time frame:
\begin{align}
    \mathcal{L}_{intv}(\mathbf{X}_{0:T}, \hat{\mathbf{X}}_{0:T}) &= \frac{1}{N * T} \sum_{n=0}^N \sum_{i=0}^T (\mathbf{x}_{n,i} * log(\hat{\mathbf{x}}_{n,i}) + (1 - \hat{\mathbf{x}_{n,i}}) * log(1 - \hat{\mathbf{x}}_{n,i}),
\end{align}
where BCE is applied to each time frame. As we are no  longer care about the solution of an initial value problem, the emphasis on $\mathbf{x}_0$ reconstruction is lifted.

\section{Experiments}
For the task of reconstructing ectopic foci that disrupt normal electrical propagation using remote observations, we include three baseline methods: \textit{ECGI}.) a first order Tikhonov regularization for reconstruction, \textit{ODE-VAE}.) the direct application of the native dynamics network to the given intervention set, and \textit{ODE-VAE-GRU}.) a Bayesian filtering ablation where a latent update mechanism (GRU) is applied at every timestep using the next 3 observations straight as input. This represents a model optimized on native and intervention dynamics together without clear disentanglement or ability to preserve \textit{intervention-free} dynamics. We denote the proposed method as \textit{ODE-VAE-IM} throughout the experiments. The observation data $\textbf{X}_{0:T}$ is passed through a Bernoulli filter between activated and deactivated nodes following min-max normalization and all network decoders have a Sigmoid activation on the output images. The intervention variable encoder (\textit{$Enc_a$}) has non-activated outputs.

Initial learning rates were found using the cyclical learning rate estimation technique \cite{smith2017cyclical} and are decayed by $0.5$ at set points throughout training. We used AdamW optimizer \cite{loshchilov2017decoupled} with a weight decay of $1e-2$, a batch size of 16, and a latent dimension of 12 across all methods. All experiments were run on NVIDIA Tesla T4s with 16GB memory, taking $\sim$8 hours to train. Our implementation and saved models are available at \url{https://github.com/qu-gg/causal-effect-neural-ssm}, along with more examples in Supplementary Material.

\textbf{Data Generation.} Following the Fitzhugh-Nagumo model \cite{izhikevich2006fitzhugh}, we simulated the transmembrane potentials on a 100*100mm 2D grid. This synthetic dataset includes two subsets 1.) native transmembrane potential set containing 1000 voltage maps, where the initial excitation location is chosen randomly across the grid and 2.) transmembrane potentials in the presence of an extra focci. Simulating the setting where the extra Foci intervenes in the normal dynamics, we generated a dataset of 705 samples with varying initial excitation location and time and location of the extra Foci.

\begin{figure}[tb]
    \centering
    \includegraphics[width=\textwidth]{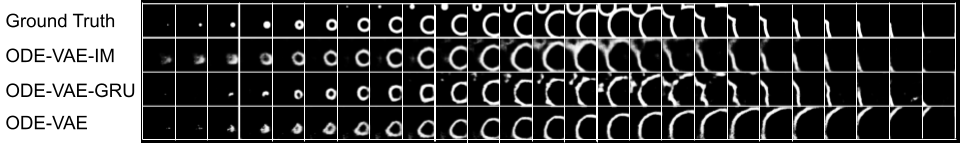}
    \caption{Synthetic TMP per-step recons in which additional excitations occur.}
    \label{fig:dualProp}
\end{figure}

\subsection{Dynamics Reconstruction}
In this section, we showcase the reconstructive and temporal capabilities of the method on normal and intervention dynamics.

\begin{figure}[tb]
    \centering
    \includegraphics[width=\textwidth]{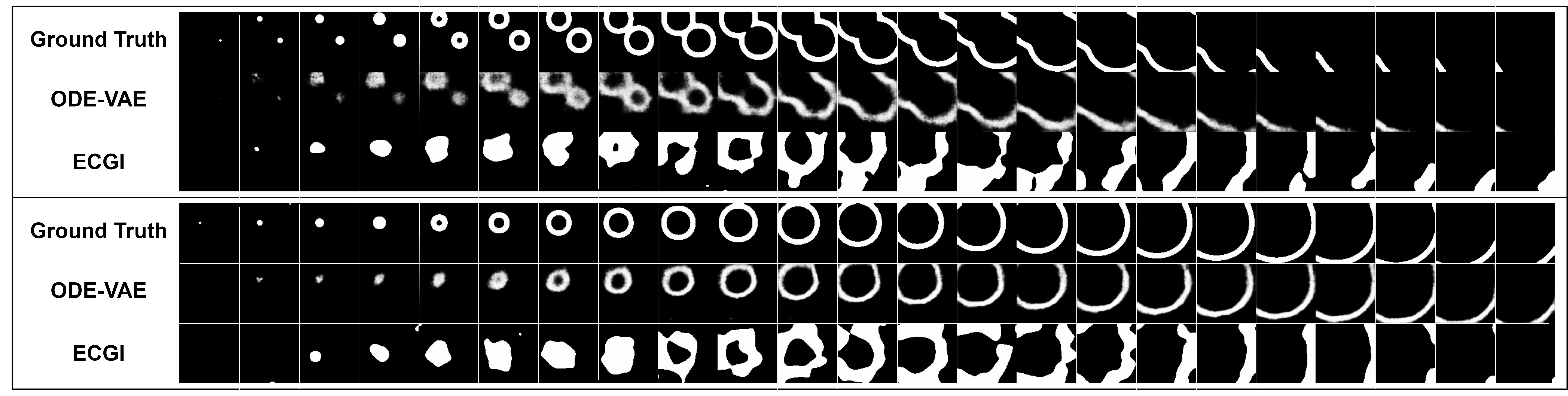}
    \caption{Reconstruction of electrical propagation in which no interventions (foci) occur.}
    \label{fig:normalProp}
\end{figure}

\textbf{Normal Dynamics.}
We first highlight the \textit{naive} dynamics model's performance on the non-intervention datasets used for pre-training. To ensure that a robust vector field is trained, our base dataset combines both single and double excitation propagations that occur within the first few frames. We posit that by introducing more propagation patterns in the base dynamics, even if they are not used in intervention, helps the ODE function and emission model to handle unstable vector state jumps. Figure~\ref{fig:normalProp} highlights two samples of per-step reconstructions, showcasing that strong dynamics are established.

\textbf{Ectopic Foci.}
Figure~\ref{fig:dualProp} highlights per-timestep reconstructions of a single sample across all methods. \textit{ODE-VAE}, the base dynamics model with no means of intervention, manages to reconstruct a propagated waveform of the \textit{native} dynamics. Leveraging this fixed native dynamics function, the presented \textit{ODE-VAE-IM} is able to uncover the hidden intervention dynamics. 

\begin{figure}[tb]
    \centering
    \includegraphics[width=\textwidth]{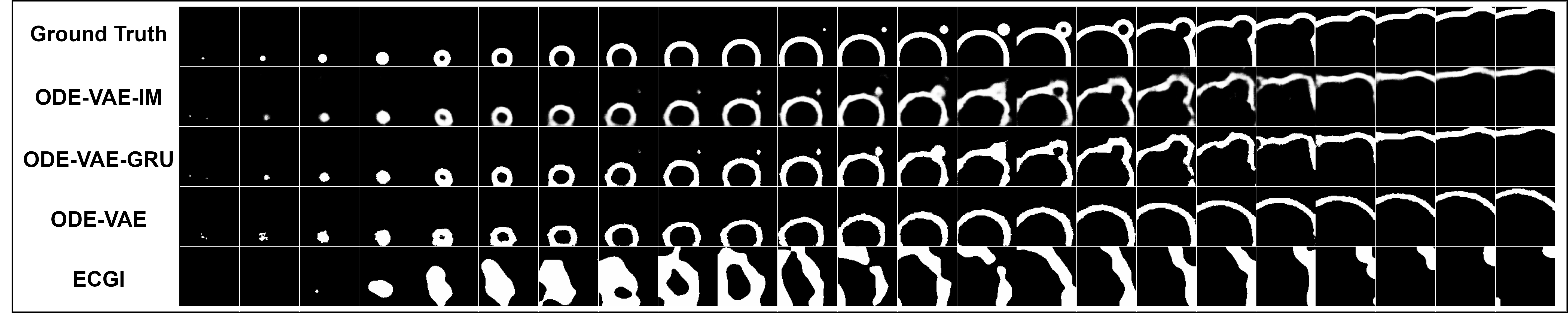}
    \caption{Reconstruction of electrical propagation in which ectopic foci occurs.}
    \label{fig:dualProp}
\end{figure}

\subsection{Localization Results}
To gauge the proposed method's numeric performance, we performed a localization test in which the time and location of interventional foci were identified from the reconstruction results across a subset of 100 samples. The mean absolute error (MAE) in time frames, the Euclidean distance in location, and the percentage of identified foci are used for quantitative metrics. Table~\ref{tab:metrics} showcases these results, in which the ability of \textit{ODE-VAE-IM} in identifying the presence of abnormal foci is highlighted.

\begin{table}[!tb]
\centering
\caption{Comparison in identifying the intervention foci's location and activation timestep. Identification percentage represents cases with clear foci reconstruction.}
\label{tab:metrics}
\begin{tabular}{|c|c|c|c|}
\hline
\textbf{Model} &  \textbf{\% Foci Identified} & \textbf{Timestep Activation MAE} & \textbf{Foci Location Error} \\
\hline
ODE-VAE-IM  &  0.85 & 3.56 & 54.90 \\
ODE-VAE-GRU &  0.79 & 3.92 & 53.71 \\
ODE-VAE     &  0.39 & 3.07 & 57.28 \\
ECGI        &  0.52 & 4.10 & 52.35 \\
\hline
\end{tabular}
\end{table}

\subsection{Latent Norm Ablation}
\begin{figure}[!tb]
    \centering
    \includegraphics[width=\textwidth]{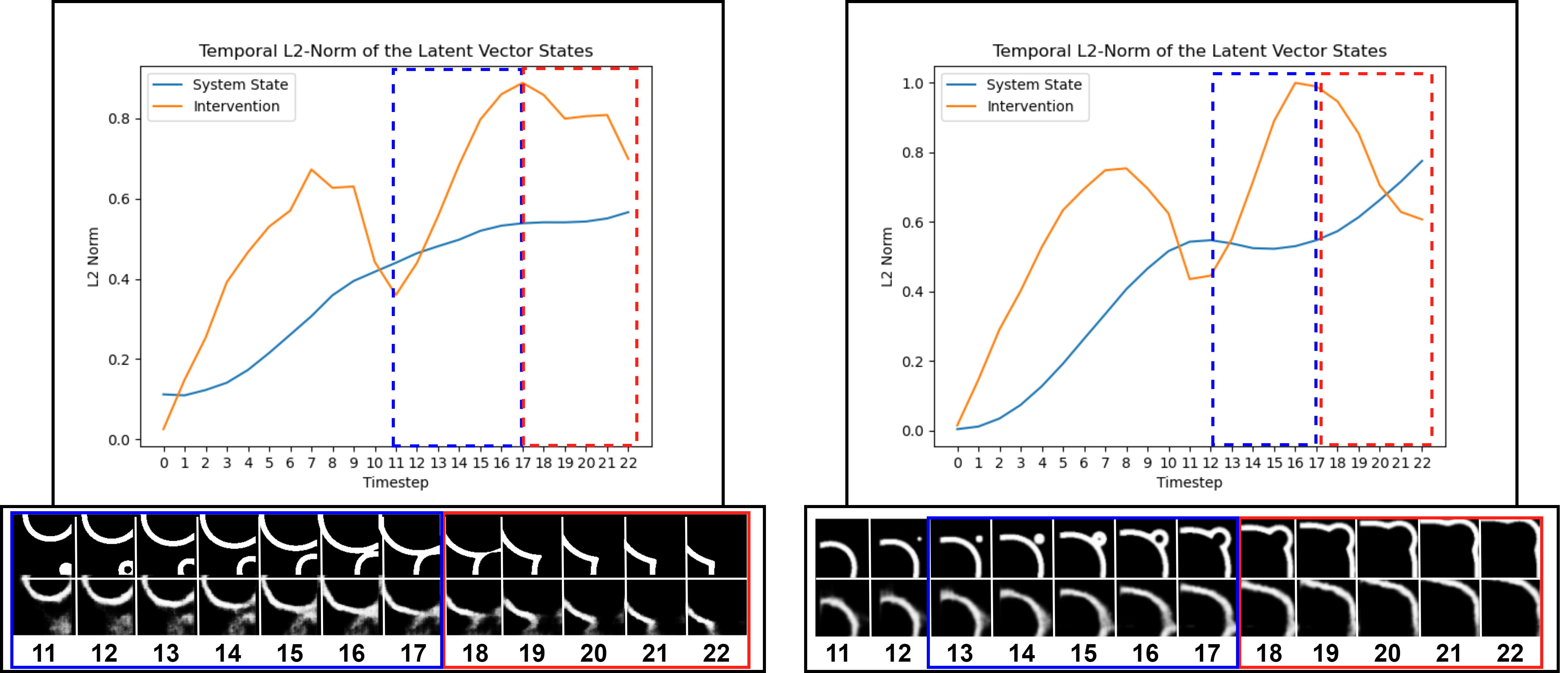}
    \caption{Visualizations of the L2-Norm of system and intervention states over time.}
    \label{fig:latentNorms}
\end{figure}

We performed an ablation study on the latent dynamics of the cause and effect by visualizing the $L^2$-Norm of their respective latent vector states over time for two examples in Figure~\ref{fig:latentNorms}. Temporal windows are highlighted in which the intervention dynamics correctly capture the windows of starting and ending intervention activity. To facilitate comparisons between each latent space, we first perform min-max normalization over the $L^2$-Norms of each latent space to bring them within the same data range. 

\section{Conclusion}
In this work, we propose a proof-of-concept interventional modelling framework to tackle the problem of low-strength local activity that signifies important abnormal triggering events in a dynamic system. We introduce a two-ODE system that separately models the cause and effect of system's latent state under the influence of hidden internal interventions. To this end we leverage pre-trained dynamical functions describing \textit{intervention-free} native dynamics of a system. We demonstrated the frameworks performance on reconstruction of the ectopic foci causing abnormality in cardiac electrical activity.

\textbf{Future Work.} Future work aims to extend this framework into 3-D inverse reconstruction and pacing localization using graph neural networks to enable experimentation on clinical data, following \cite{jiang2020learning,jiang2021label}. One promising direction is learning multi-interventional dynamics simultaneously via attention mechanisms.

\textbf{Limitations.} The transition between training $\mathcal{F}_{z}$ and $\mathcal{F}_{a}$ has a period of re-training up to the original performance, which identity-based initialization or loss-influence annealing $\mathcal{F}_{a}$ may alleviate. $Enc_z$ has potential generalization problems when the initial frame distributions shift between native and intervention sets and requires experimentation on its general training setup.

\bibliographystyle{splncs04}
\bibliography{Ref}
\end{document}